%% file: FormalizingPiecewiseAffineFunctionsinCoq.tex
\documentclass[runningheads]{llncs}
\usepackage[T1]{fontenc}
\usepackage{graphicx}
\usepackage{todonotes}
\usepackage{amsmath}
\usepackage{amssymb}
\usepackage{cite}

\usepackage[usestackEOL]{stackengine}
\stackMath

\input{preambel}

\begin{document}
\title{Formalizing Piecewise Affine Activation Functions of Neural Networks in \coq}
\titlerunning{Piecewise Affine Activation Functions of Neural Networks in \coq}
% If the paper title is too long for the running head, you can set
% an abbreviated paper title here
%
\author{Andrei~Aleksandrov\orcidID{0000-0002-4717-4206} \and
Kim~V\"ollinger\orcidID{0000-0002-8988-0053}}
\authorrunning{A. Aleksandrov \& K. V\"ollinger}
% First names are abbreviated in the running head.
% If there are more than two authors, 'et al.' is used.
%
\institute{Technische Universit\"at Berlin, Germany\\
\email{andrei.aleksandrov@campus.tu-berlin.de}\\ 
\email{voellinger@tu-berlin.de}}
\maketitle              % typeset the header of the contribution
\begin{abstract}
Verification of neural networks relies on activation functions being \emph{piecewise affine} (\pwa) ---
enabling an encoding of the verification problem for theorem provers.
In this paper, we present the first formalization of \pwa activation functions 
for an interactive theorem prover tailored to verifying neural networks within \coq using the library \coquelicot for real analysis.
As a proof-of-concept, we construct the popular \pwa activation function \relu.
We integrate our formalization into a \coq model of neural networks, 
and devise a verified transformation from a neural network $\mathcal{N}$ to a \pwa function  representing $\mathcal{N}$
by composing \pwa functions that we construct for each layer.
This representation enables encodings for proof automation, e.g. \coq's tactic \texttt{lra} -- a decision procedure for linear real arithmetic.
Further, our formalization paves the way for integrating \coq in frameworks of neural network verification
as a fallback prover when automated proving fails.

\end{abstract}

\input{sections/introduction.tex}
%\input{sections/related-work.tex}
\input{sections/preliminaries.tex}

%main parts
\input{sections/pwa-fcts-in-coq.tex}
\input{sections/nn-as-pwa-in-coq.tex}
\input{sections/discussion.tex}

\bibliographystyle{splncs04}
\bibliography{mybibliography}
\end{document}

%% file: preambel.tex
\usepackage{xspace}
\newcommand{\coq}{\mbox{\textsc{Coq}}\xspace} 
\newcommand{\coquelicot}{\mbox{\textsc{Coquelicot}}\xspace} 
\newcommand{\pwa}{\mbox{\textsc{pwa}}\xspace} 
\newcommand{\relu}{\mbox{\textsc{ReLU}}\xspace} 
\newcommand{\linear}{\mbox{\textsc{Linear}}\xspace} 
\newcommand{\R}{\mathbb{R}} 
\newcommand{\Pset}{\mathbf{P}}
\newcommand{\Aset}{\mathbf{A}}
\newcommand{\N}{\mathcal{N}}
\newcommand{\C}{\mathcal{C}}

\usepackage{float}
\usepackage{listings} 
\definecolor{dkviolet}{rgb}{0.4,0,0.4}
\definecolor{dkgreen}{rgb}{0,0.6,0}
\definecolor{ltblue}{rgb}{0,0,0.35}
\definecolor{dkblue}{rgb}{0,0,0.6}
\definecolor{dkred}{rgb}{0.6,0,0}
\definecolor{ltgray}{rgb}{0.95,0.95,0.95}

\input{coq.sty}

%% file: sections/introduction.tex
\section{Introduction}
\label{s:introduction}

%research area 
The growing importance of neural networks motivates the search of verification techniques for them.
Verification with \emph{automatic} theorem provers is vastly under study,
usually targeting feedforward networks with \emph{piecewise affine} (\pwa) activation functions since
the verification problem can be then encoded as an \textsc{SMT} or \textsc{MILP} problem.
In contrast, few attempts exist on investigating \emph{interactive} provers.
Setting them up for this task though offers not only a fallback option when automated proving fails
but also insight on the verification process.

That is why in this paper, we work towards this goal by presenting the first formalization of \pwa activation functions 
for an interactive theorem prover tailored to verifying neural networks with \coq.
We constructively define \pwa functions using the polyhedral subdivision of a \pwa function~\cite{Scholtes2012}
since many algorithms working on polyhedra are known~\cite{combinatorial-optimization}
with some tailored to reasoning about reachability properties of neural networks~\cite{rpm2021}.  
Motivated by verification, we restrict \pwa functions by a polyhedron's constraint to be \emph{non-strict}
in order to suit linear programming~\cite{vanderbei2020linear} and by employing \emph{finitely} many polyhedra to fit \textsc{SMT}/\textsc{MILP} solvers~\cite{vanderbei2020linear, de2011satisfiability}.
We use reals supported by the library \coquelicot to enable reasoning about gradients and matrices with \coq's standard library providing the tactic \texttt{lra} -- a decision procedure for linear real arithmetic.
As a proof-of-concept, we construct the activation function \relu -- one of the most popular in industry~\cite{FundamentalsNN} and formal verification~\cite{10.5555/3327345.3327388}.
Furthermore, we devise a sequential \coq model of feedforward neural networks integrating \pwa activation layers. 
Most importantly, we present a verified transformation from a neural network $\mathcal{N}$ to a \pwa function $f_\mathcal{N}$ representing $\mathcal{N}$
with the main benefit being again encodings for proof automation.
To this end, we introduce two verified binary operations on \pwa functions -- usual function composition and
an operator to construct a \pwa function for each layer.
In particular, we provide the following contributions with the corresponding \coq code available 
on \textsc{GitHub}\footnote{At \url{https://github.com/verinncoq/formalizing-pwa} with matrix\_extensions.v (Section~\ref{s:preliminaries}), piecewise\_affine.v (Section~\ref{ss:pwa}), neuron\_functions.v (Section~\ref{ss:relu}), neural\_networks.v (Section~\ref{ss:nn-model} and~\ref{ss:transformation}) and pwaf\_operations.v (Section~\ref{ss:composition} and~\ref{ss:concatenation}).}:
\begin{enumerate}
 \item a formalization of \pwa functions based on polyhedral subdivision tailored to verification of neural networks (Section~\ref{s:pwa-fcts-in-coq}),
 \item a construction of the popular activation function \relu (Section~\ref{s:pwa-fcts-in-coq}),
 \item a sequential model for feedforward neural networks with parameterized layers (Section~\ref{s:nn-as-pwa-in-coq}),
  \item composition for \pwa functions and an operator for constructing higher dimensional \pwa functions out of lower dimensional ones (Section~\ref{s:nn-as-pwa-in-coq}), and
 \item a verified transformation from a feedforward neural network with \pwa activation to a single \pwa function representing the network (Section~\ref{s:nn-as-pwa-in-coq}).
\end{enumerate}

\paragraph{Related Work.}
A variety of work on using automatic theorem provers to verify neural networks exists with 
the vast majority targeting feedforward neural networks with \pwa activation functions~\cite{10.1561/2400000035,10.5555/3327345.3327388,imagestar,VerificationLP,Scheibler2015TowardsVO,Ehlers2017FormalVO,botoeva}.
In comparison, little has been done regarding interactive theorem provers with some 
mechanized results from machine learning~\cite{10.1145/3088525.3088673,expressiveness-isabelle},
a result on verified training in \textsc{Lean}~\cite{10.5555/3305890.3305996}
and, relevant to this paper,  pioneering work on verifying networks in \textsc{Isabelle}~\cite{isabelle-nn}
and in \coq~\cite{Bagnall2019CertifyingTT}.
Apart from~\cite{isabelle-nn} targeting \textsc{Isabelle} instead of \coq, 
both network models are not generalized by entailing a formalization of \pwa functions
and in addition they do not offer a model of the network as a (\pwa) function -- both contributions of this paper.
%Not directly related but noteworthy are the efforts on formalizing tensors in \coq since they play into 
%lying foundations on neural network verification in \coq as worked towards to in this paper.

%% file: sections/preliminaries.tex
%expected to be small
%probably part of introduction
\section{Preliminaries}
\label{s:preliminaries}

We clarify notations and definitions important to this paper. 
We write $dom(f)$ for a function's domain, $dim(f)$ for the dimension of $dom(f)$ and $(f \circ g)(x)$ for function composition. 
For a matrix $M$, $M^T$ is the transposed matrix. 
We consider block matrices. To clarify notation, consider a block matrix made out of matrices $M_1, ..., M_4$:
$$\begin{bmatrix}
     M_1 & \vline & M_2 \\
     \hline 
     M_3 & \vline & M_4
    \end{bmatrix}$$
\subsection{Piecewise Affine Topology}
We give the important definitions regarding \pwa functions~\cite{Rourke1982IntroductionTP, ziegler1995lectures, Scholtes2012}.

\begin{definition}[Linear Constraint]
For some $c \in \R^n, b \in \R$, a \emph{linear constraint} is an inequality of form $c^Tx\leq b$ for any $x \in \R^n$. 
\end{definition}

\begin{definition}[Polyhedron\footnote{In the literature often referred to as a convex, closed polyhedron.}]
A \emph{polyhedron} $P$ is the intersection of finitely many halfspaces, meaning $P := \{x \in \R^n |  c_1^Tx\leq b_1 \land ... \land c_m^Tx\leq b_m\}$ with $c_i, b_i \in \R^n, b_i \in \R$ and $i \in \{1, ..., m\}$. 
\end{definition}
We denote the constraints of $P$ as $\C(P) := \{(c_1^Tx\leq b_1), ..., (c_m^Tx\leq b_m)\}$ for readability even though a constraint is given by $c_i$ and $b_i$ while $x$ is arbitrary.

\begin{definition}[Affine Function\footnote{A linear function is a special case of an affine function \cite{YANG201919}. However, in literature, the term linear is sometimes used for both.}]
A function $f: \R^m \rightarrow \R^n$ is called \emph{affine} if there exists $M \in \R^{n \times m}$ and $b \in \R^n$ such that for all $x \in \R^m$ holds: $f(x) = Mx + b$.
\end{definition}
 
\begin{definition}[Polyhedral Subdivision]
A \emph{polyhedral subdivision} $S \subseteq \R^n$ is a finite set of polyhedra $\Pset := \{P_1, \dots, P_m\}$ such that (1) $S = \bigcup_{i = 1}^m P_i$ and (2) for all $P_i, P_j \in \Pset, x \in P_i \cap P_j$, and for all $\epsilon > 0$ there exists $x'$ such that $|x - x'| < \epsilon$, and $x' \notin P_i \cap P_j$.
\end{definition}

\begin{definition}[Piecewise Affine Function]
A continuous function $f: D \subseteq \R^m \rightarrow \R^n$ is \emph{piecewise-affine} if there is a polyhedral subdivision $\Pset = \{P_1,\dots,P_l\}$ of $D$ and a set of affine functions $\{f_1,\dots,f_l\}$ such that for all $x \in D$ holds $f(x) = f_i(x)$ if $x \in P_i$.
\end{definition}

\subsection{Neural Networks}

Neural networks approximate functions by learning from sample points during training~\cite{calin2020deep} with arbitrary precision~\cite{cybenko1989approximation, hornik1991approximation, math7100992}. 
A feedforward neural network is a directed acyclic graph with the edges having weights and the vertices (neurons) having biases and being structured in layers.
Each layer applies a generic affine function for summation and an activation function (possibly a \pwa function). 
In many machine learning frameworks (e.g. \textsc{PyTorch}), these functions are modelled as separate layers followed up by each other.
We adopt this structure in our \coq model with a \textit{linear layer} implementing the generic affine function. 
Every network has an input and an output layer with optional hidden layers in between.

\subsection{Interactive Theorem Prover \coq \& Library \coquelicot}

We use the interactive theorem prover \coq~\cite{coqref} providing a non-turing-complete functional programming language extractable to selected functional programming languages  
 and a proof development system -- 
a popular choice for formal verification of programs and formalization of mathematical foundations. 
Additionally, we use the real analysis library \coquelicot~\cite{boldo2015coquelicot} offering derivatives, integrals, and matrices compatible 
with \coq's standard library.

\paragraph{Extensions in \coq: Column Vectors \& Block Matrices.}
For this paper, we formalized column vectors and block matrices on top of \coquelicot. 
A column vector \texttt{colvec} is identified with matrices and equipped with a dot product~\texttt{dot} on vectors and some additional lemmas to simplify proofs.
Additionally, we formalized several notions for \coquelicot's matrix type. We provide multiplication of a matrix with a scalar \texttt{scalar\_mult} and
transposition \texttt{transpose} of matrices. We provide operations on different shapes of matrices and vectors such as
a right-to-left construction of block diagonal matrices \texttt{block\_diag\_matrix}, a specialization thereof on vectors \texttt{colvec\_concat} and
extensions of vectors with zeroes on the bottom \texttt{extend\_colvec\_at\_bottom} or top \texttt{extend\_colvec\_on\_top}, denoted as follows:
$\begin{bmatrix}
     M_1 & \vline & 0 \\
     \hline 
     0 & \vline & M_2
    \end{bmatrix},
\begin{bmatrix}
        \vec v_1 \\
        \hline
        \vec v_2
    \end{bmatrix},
\begin{bmatrix}
        \vec v \\
        \hline
        \vec 0
    \end{bmatrix},\text{ and}
\begin{bmatrix}
        \vec 0 \\
        \hline
        \vec v
    \end{bmatrix}\text{.}     
$
Moreover, we proved lemmas relating all new operations with each other and the existing matrix operations.

%% file: sections/pwa-fcts-in-coq.tex
\section{Formalization of Piecewise Affine Functions in Coq}
\label{s:pwa-fcts-in-coq}

We formalize \pwa functions tailored to neural network verification with \pwa activation. As a proof-of-concept, we construct the activation function Rectified Linear Unit (\relu) -- one of the most popular activation functions in industry~\cite{FundamentalsNN} and formal verification~\cite{10.5555/3327345.3327388}. 

%%%%%%%%%%%%%%%%%%%%%%%%%%%%%%%%%%%%%%%%%%%%%%%%%%%%%%%%%%%%%%%%%%%%%%%%%%
\subsection{Inductive Definition of PWA Functions}
\label{ss:pwa}

We define a linear constraint with a dimension \emph{dim} and parameters, vector $c \in \mathbb{R}^{dim}$ and scalar $b \in \mathbb{R}$,
being satisfied for a vector $x \in \mathbb{R}^{dim}$ if 
$c \cdot x \leq b$:
\begin{lstlisting}
Inductive LinearConstraint (dim:nat) : Type := 
| Constraint (c: colvec dim)(b: R).

Definition satisfies_lc {dim: nat} (x: colvec dim) (l: LinearConstraint dim)
: Prop := match l with | Constraint c b => dot c x <= b end. 
\end{lstlisting}

We define a polehydron as a finite set of linear constraints together with a predicate stating that a point lies in a polyhedron:
\begin{lstlisting}
Inductive ConvexPolyhedron (dim: nat) : Type :=
| Polyhedron (constraints: list (LinearConstraint dim)).
Definition in_convex_polyhedron {dim: nat} (x: colvec dim) (p: ConvexPolyhedron dim) :=
match p with | Polyhedron lcs =>
  forall constraint, In constraint lcs -> satisfies_lc x constraint end.
\end{lstlisting}

Finally, we define a \pwa function as a record composed of the fields \texttt{body} holding the polyhedral subdivision for piecewise construction of the function, 
and \texttt{prop} for the property univalence (i.e. all ``pieces'' together yield a function).
\begin{lstlisting}
Record PWAF (in_dim out_dim: nat): Type := mkPLF {
    body: list (ConvexPolyhedron in_dim * ((matrix out_dim in_dim) * colvec out_dim));
    prop: pwaf_univalence body; }.
\end{lstlisting}

\paragraph{Piecewise Construction.}
We construct a \pwa function $f$ by a list of polyhedra, matrices and vectors with a triple $(P,M,b )$ defining a ``piece'' of $f$
by an affine function with $f(x) = Mx + b$  if $x \in P$.
For evaluation, we search a polyhedron containing \textit{x} and compute the affine function:
\begin{lstlisting}
Fixpoint pwaf_eval_helper {in_dim out_dim: nat}
    (body: list (ConvexPolyhedron in_dim * ((matrix (T:=R) out_dim in_dim) * colvec out_dim))) (x: colvec in_dim) 
    : option (ConvexPolyhedron in_dim * ((matrix out_dim in_dim) * colvec out_dim)) :=
    match body with
    | nil => None
    | body_el :: next => 
        match body_el with
        | (polyh, (M, b)) =>
            match polyhedron_eval x polyh with
            | true => Some (body_el)
            | false => pwaf_eval_helper next x
    end end end.
\end{lstlisting}
To handle the edge case where no such polyhedron is found (i.e. $x \notin dom(f)$), we use a wrapper function \texttt{pwaf\_eval}.
For the purpose of proving, we define a predicate \texttt{in\_pwaf\_domain} for the existence of such a polyhedron 
and a predicate \texttt{is\_pwaf\_value} for a stating the function is evaluated to a certain value.

\paragraph{Univalence.}
We enforce the construction to be a function by stating univalence, in this case all pairs of polyhedra having coinciding affine functions in their intersection,
requiring a proof for each instance of type \texttt{PWAF}:
\begin{lstlisting}
Definition pwaf_univalence {in_dim out_dim: nat}
    (l: list (ConvexPolyhedron in_dim * 
      ((matrix out_dim in_dim) * colvec out_dim))) :=
    ForallPairs (fun e1 e2 => let p1 := fst e1 in let p2 := fst e2 in
        forall x, in_convex_polyhedron x p1 /\ in_convex_polyhedron x p2 ->
          let M1 := fst (snd e1) in let b1 := snd (snd e1) in
          let M2 := fst (snd e2) in let b2 := snd (snd e2) in
            Mplus (Mmult M1 x) b1 = Mplus (Mmult M2 x) b2 ) l. 
\end{lstlisting}

\paragraph{Class of Formalized PWA Functions.}
Motivated by \pwa activation functions in the context of neural network verification, 
our \pwa functions are restricted by 
\begin{itemize}
\item[(1)] all linear constraints being \emph{non-strict}, and 
\item[(2)] being defined over a union of \emph{finitely} many polyhedra.
\end{itemize}
Restriction (1) is motivated by linear programming usually dealing with non-strict constraints~\cite{vanderbei2020linear}, and restriction (2) by MILP/SMT solvers commonly accepting finitely many variables~\cite{vanderbei2020linear, de2011satisfiability}. Since we use that every continuous \pwa function on $\R^n$ admits a polyhedral subdivision of the domain~\cite{Scholtes2012}, all continuous \pwa functions with a finite subdivision can be encoded.  

For \pwa functions not belonging to this class, consider any discontinuous \pwa function since discontinuity violates restriction (1), 
and any periodic \pwa function as excluded by restriction (2) due to having infinitely many "pieces".

\paragraph{Choice of Formalization.}
We use real numbers (instead of e.g. rationals or floats) to enable \coquelicot's reasoning about derivatives interesting for neural networks' gradients. 
\coquelicot builds up on the reals of \coq's standard library allowing the use of \coq's tactic \texttt{lra} -- a \coq-native decision procedure for linear real arithmetic.

Moreover, we use inductive types since they come with an induction principle and therefore ease proving. In addition, the type \texttt{list} (e.g. used for the definition of \pwa functions) enjoys extensive support in \coq. For example, \texttt{pwaf\_univalence} is stated using 
the list predicate \texttt{ForAllPairs} and proofs intensively involve lemmas from \coq's standard library.

A constructive definition using the polyhedral subdivision is interesting since many efficient algorithms are known that work on polyhedra~\cite{combinatorial-optimization}
with even algorithms tailored to neural network verification around~\cite{rpm2021}.
We expect that such algorithms are implementable in an idiomatic functional style using our model.
Furthermore, we anticipate easy-to-implement encodings for proof automation.

%%%%%%%%%%%%%%%%%%%%%%%%%%%%%%%%%%%%%%%%%%%%%%%%%%%%%%%%%%%%%%%%%%%%%%%%%%
\subsection{Example: Rectified Linear Unit Activation Function}
\label{ss:relu}
We construct \relu as a \pwa function defined by two ``pieces'' each of which being a linear function. The function is defined as:
$$\relu(x) := 
\begin{cases}
    0, & x < 0 \\
    x, & x \geq 0
\end{cases}$$
\paragraph{Piecewise Construction.}
The intervals, $(- \infty, 0)$ and $[0, \infty)$, each correspond to a polyhedron in $\R$ defined by a single constraint\footnote{Matrices involved are one-dimensional vectors since \relu is one-dimensional. For technical reasons, in \coq, the spaces $\R$ and $\R^1$ differ with the latter working on one-dimensional vectors instead on scalars.}:
$P_{left} := \{x \in \R^1 | [1] \cdot x <= 0\}$ and $ P_{right} := \{x \in \R^1 | [-1] \cdot x <= 0 \}$.
We define these polyhedra as follows:\footnote{\texttt{Mone} is \coquelicot's identity matrix which in this case is a one-dimensional vector.
\texttt{scalar\_mult} is multiplication of a matrix by a scalar (see Section~\ref{s:preliminaries}).}
\begin{lstlisting}
Definition ReLU1d_polyhedra_left := Polyhedron 1 [Constraint 1 Mone 0].
Definition ReLU1d_polyhedra_right 
    := Polyhedron 1 [Constraint 1 (scalar_mult (-1) Mone) 0].
\end{lstlisting}

\relu's construction list contains these polyhedra each associated with a matrix  
and vector, in these cases $([0],[0])$ and $([1],[0])$, for the affine functions:
\begin{lstlisting}
Definition ReLU1d_body: list (ConvexPolyhedron 1 * (matrix (T:=R) 1 1 * colvec 1)) 
    := [(ReLU1d_polyhedra_left, (Mzero, null_vector 1));
        (ReLU1d_polyhedra_right, (Mone, null_vector 1))].
\end{lstlisting}

\paragraph{Univalence.}
Note that while \relu's intervals are distinct, the according polyhedra with non-strict constraints are not. 
To ensure the construction to be a function, we prove univalence by proving that only $ [0] \in (P_{left} \cap P_{right})$:
\begin{lstlisting}
Lemma RelU1d_polyhedra_intersect_0:
    forall x, in_convex_polyhedron x ReLU1d_polyhedra_left /\ 
      in_convex_polyhedron x ReLU1d_polyhedra_right -> x = null_vector 1.
\end{lstlisting}
Finally, we ensure for each polyhedra pair holds $[1] \cdot [0] + [0] = [0] \cdot [0] + [0]$,
and instantiate a \texttt{PWAF} by
\lstinline{Definition ReLU1dPWAF := mkPLF 1 1 ReLU1d_body ReLU1d_pwaf_univalence.}

\paragraph{On the Construction of \pwa Functions.}
Analogously to the \relu example, other activation functions sharing its features of being one-dimensional and consisting of a few polyhedra can be constructed similarly. 
We can construct a multi-dimensional version out of a one-dimensional function as we will illustrate for \relu in Section~\ref{ss:concatenation}.
Activation functions that require more effort to construct are for example different types of pooling~\cite{calin2020deep}, mostly due to a non-trivial polyhedra structure and inherent multi-dimensionality. This effort motivates future development of more support in constructing \pwa functions with the goal to compile a library of layer types.

%% file: sections/nn-as-pwa-in-coq.tex
\section{Verified Transformation of a Neural Network to a \pwa Function}
\label{s:nn-as-pwa-in-coq}

We present our main contribution: a formally verified transformation of a feedforward neural network with \pwa activations into a single \pwa function.
First, we introduce a \coq model for feedforward neural networks (Section~\ref{ss:nn-model}).
We follow up with two verified binary operations on \pwa functions at the heart of the transformation, \textit{composition} (Section~\ref{ss:composition}) and \textit{concatenation} (Section~\ref{ss:concatenation}),
and finish with the verified transformation (Section~{\ref{ss:transformation}}).

\subsection{Neural Network Model in \textsc{Coq}}
\label{ss:nn-model}

We define a neural network \textit{NNSequential} as a list-like structure containing layers parameterized on the type of activation, and the input's, output's and hidden layer's dimensions with dependent types preventing dimension mismatch:
\begin{lstlisting}
Inductive NNSequential {input_dim output_dim: nat} :=
| NNOutput : NNSequential
| NNPlainLayer {hidden_dim: nat}:
    (colvec input_dim -> colvec hidden_dim)
    -> NNSequential (input_dim:=hidden_dim) (output_dim:=output_dim)
    -> NNSequential
| NNPWALayer {hidden_dim: nat}:
    PWAF input_dim hidden_dim 
    -> NNSequential (input_dim:=hidden_dim) (output_dim:=output_dim)
    -> NNSequential
| NNUnknownLayer {hidden_dim: nat}:
    NNSequential (input_dim:=hidden_dim) (output_dim:=output_dim)
    -> NNSequential.
\end{lstlisting}
The network model has four layer types:
\texttt{NNOutput} as the last layer propagates input values to the output; 
\texttt{NNPlainLayer} is a layer allowing any function in \coq defined on real vectors;
\texttt{NNPWALayer} is a \pwa activation layer -- the primary target of our transformation; and
\texttt{NNUnknownLayer} is a stub for a layer with an unknown function.

Informally speaking, the semantics of our model is as follows: 
for a layer \texttt{NNOutput} the identity function\footnote{We use the customized identity function \textit{flex\_dim\_copy}.} is evaluated, 
for \texttt{NNPlainLayer} the passed function, 
for \texttt{NNPWALayer} the passed \pwa function, 
and for \texttt{NNUnknownLayer} a failure is raised.
Thus, the \textit{NNSequential} type does not prescribe any specific functions of layers but expects them as parameters.

\paragraph{An Example of a Neural Network.}

In order to give an example, we define specific layers for a network, in this case the \pwa layers \linear and \relu:
\begin{lstlisting}
Definition NNLinear {input_dim hidden_dim output_dim: nat} 
  (M: matrix hidden_dim input_dim) (b: colvec hidden_dim) 
  (NNnext: NNSequential (input_dim:=hidden_dim) (output_dim:=output_dim)) 
  := NNPWALayer (LinearPWAF M b) NNnext.

Definition NNReLU {input_dim output_dim: nat} 
  (NNnext: NNSequential (input_dim:=input_dim) (output_dim:=output_dim)) 
  := NNPWALayer (input_dim:=input_dim) ReLU_PWAF NNnext.    
\end{lstlisting}

%Unknown layers lead to an evaluation failure
%Plain layers are evaluated directly using a function provided as a parameter
% PWAF layers utilize \textit{PWAF\_eval} described in Section \ref{s:pwa-fcts-in-coq}, but may also lead to evaluation failure in case layer input is not in a domain of the PWA function of the layer
% Output layers processed using custom identity function \textit{flex\_dim\_copy}. Limitation of our definition is that input and output dimensions of the output layer are not necessarily equal, which is handled by dropping some inputs in case when input dimension is greater than the output dimension or by filling additional outputs with zeroes.

%The evaluation of a neural network is implemented in \coq as follows: 
%\begin{lstlisting}
% Fixpoint nn_eval {in_dim out_dim: nat} 
%     (nn: NNSequential (input_dim:=in_dim) (output_dim:=out_dim)) 
%     (input: colvec in_dim): option (colvec out_dim)
%     := 
%     match nn with
%         | NNOutput => Some (flex_dim_copy input)
%         | NNPlainLayer _ f next_layer => 
%             nn_eval next_layer (f input)
%         | NNPWALayer _ pwaf next_layer => 
%             match pwaf_eval pwaf input with
%             | Some output => nn_eval next_layer output
%             | None => None
%             end  
%         | NNUnknownLayer _ _ => 
%             None
%     end.
%\end{lstlisting}

As an example, we consider a neural network with these two layers:
\begin{lstlisting}
Definition example_weights: matrix 2 2 := [[2.7, 0],[1, 0.01]].
Definition example_biases: colvec 2 := [[1], [0.25]].
Definition example_nn := (NNLinear example_weights example_biases 
                         (NNReLU (NNOutput (output_dim:=2)))).
\end{lstlisting}

\paragraph{From a Trained Neural Network into the World of \coq.}
As illustrated, we can construct feedforward neural networks in \coq. 
Another option is to convert a neuronal network trained outside of \coq into an instance of the model.
In~\cite{Bagnall2019CertifyingTT} a python script is used for conversion from \textsc{PyTorch} to their \coq model without any correctness guarantess,
while in~\cite{isabelle-nn} an import mechanism from \textsc{TensorFlow} into \textsc{Isabelle} is used, where correctness of the import has to be established for each instance of their model. 
We are working with a converter expecting a neural network in the \textsc{ONNX} format 
(i.e. a format for neural network exchange supported by most frameworks)~\cite{onnx}
to produce an according instance in our \coq model~\cite{GummersbachBA23}.\footnote{A bachlor thesis supervised by one of the authors and scheduled for publication.}
This converter is mostly written within \coq with its core functionality being verified. 

\paragraph{Choice of Model.}
While feedforward neural networks are often modeled as directed acyclic graphs~\cite{Aggarwal2021, Kruse2022}, 
in the widely used machine learning frameworks \textsc{TensorFlow} and \textsc{PyTorch}
a sequential model of layers is employed as well. Our model corresponds to the latter, and is inspired by the, to our knowledge, only published neural network model in \coq (having been used for generalization proofs)~\cite{Bagnall2019CertifyingTT}.
Our model though is more generic by having parameterized layers instead of being restricted to \relu activation. Moreover, while their model works with customized floats,
we decided for reals in order to support \coquelicot's real analysis as discussed in Section~\ref{s:pwa-fcts-in-coq}.

A graph-based model carries the potential to be extended to other types of neural networks such as recurrent networks featuring loops in the length of the input.
For the reason of being generic, \textsc{ONNX} employs a graph-based model.
Hence, an even more generic graph-based \coq model is in principle desirable but it also adds complexity.
In~\cite{isabelle-nn} the focus is on a sequential model which the authors showed to be superior to a graph-based model for the purpose of verification.
Hence, we expect that the need for a sequential \coq model for \emph{feedforward} networks will stay even in the existence of a graph-based model.

%%%%%%%%%%%%%%%%%%%%%%%%%%%%%%%%%%%%%%%%%%%%%%%%%%%%%%%%%%%%
\subsection{Composition of PWA functions}
\label{ss:composition}

Besides composition being a general purpose binary operation closed over \pwa functions~\cite{Scholtes2012}, 
it is needed in our transformation  to compose \pwa layers.
Since for \pwa functions $f: \mathbb{R}^l \rightarrow \mathbb{R}^n$ and $g: \mathbb{R}^m \rightarrow \mathbb{R}^l$ their composition $z = f \circ g$ is a \pwa function,
composition in \coq produces an instance of type \texttt{PWAF} requiring a construction and a proof of univalence:
\begin{lstlisting}
Definition pwaf_compose {in_dim hidden_dim out_dim: nat} 
    (f: PWAF hidden_dim out_dim) (g: PWAF in_dim hidden_dim)
    : PWAF in_dim out_dim := mkPLF in_dim out_dim 
        (pwaf_compose_body f g) (pwaf_compose_univalence f g).
\end{lstlisting}

\paragraph{Piecewise Construction of Composition.}
Assume a \pwa function $f$ defined on the polyhedra set $\Pset^f = \{P^f_1, \dots, P^f_k\}$  with affine functions given by the parameter set 
$\Aset^f = \{(M^f_1, b^f_1), \dots, (M^f_K, b^f_k)\}$. Analogously, $g$ is given by $\Pset^g$ and $\Aset^g$.  
For computing a composed function $z = f \circ g$ at any $x \in \mathbb{R}^m$, we need a polyhedron $P^g_j \in \Pset^g$ such that $x \in P^g_j$ to compute $g(x) = M^g_jx + b^g_j$ 
with $(M^g_j, b^g_j) \in \Aset^g$. Following, we need a polyhedron $P^f_i \in \Pset^f$ with $g(x) \in P^f_i$ to finally compute $z(x) = M^f_ig(x) + b^f_i$ with $(M^f_i, b^f_i) \in \Aset^f$. 

We have to reckon on function composition on the level of polyhedra sets to construct $z$'s polyhedra set $\Pset^z$.
%as denoted by $\Pset^z = \Pset^f \circ \Pset^g$.
For each pair $P^f_i \in \Pset^f$, $P^g_j \in \Pset^g$, we create a polyhedron $P^z_{i,j} \in \Pset^z$ such that
$x \in P^z_{i,j}\text{ iff }x \in P^g_j\text{ and }M^g_jx + b^g_j \in P^f_i$ with $(M^g_j, b^g_j) \in \Aset^g$. 
Consequently, $\C(P^g_j) \subseteq \C(P^z_{i,j})$ while the constraints of $P^f_i$ have to be modified. 
For $(c_i \cdot x \leq b_i) \in \C(P^f_i)$ we have the modified constraint $((c_i^TM^g_j) \cdot x \leq b_i - c_i \cdot b^g_j) \in \C(P^z_{i,j})$. 
We construct a polyhedra set accordingly in \coq including empty polyhedra in case no qualifying pair of polyhedra exists:
\begin{lstlisting}
Fixpoint compose_polyhedra_helper {in_dim hidden_dim: nat} 
    (M: matrix hidden_dim in_dim) (b1: colvec hidden_dim)
    (l_f: list (LinearConstraint hidden_dim)) :=
    match l_f with 
    | [] => []
    | (Constraint c b2) :: tail =>
        Constraint in_dim 
            (transpose (Mmult (transpose c) M)) (b2 - (dot c b1)) :: 
                compose_polyhedra_helper M b1 tail
    end.
        
Definition compose_polyhedra {in_dim hidden_dim: nat} 
    (p_g: ConvexPolyhedron in_dim) 
    (M: matrix hidden_dim in_dim) (b: colvec hidden_dim)
    (p_f: ConvexPolyhedron hidden_dim) :=
    match p_g with | Polyhedron l1 =>
        match p_f with | Polyhedron l2 => 
            Polyhedron in_dim (l1 ++ compose_polyhedra_helper M b l2)
    end end.
\end{lstlisting}

Further, each $(M^z_{i,j}, b^z_{i,j}) \in \Aset^z$ is defined as $(M^f_jM^g_i, M^f_jb^g_i + b^f_j)$ as a result of usual composition of two affine functions:
% we apply the usual function composition. Let us examine a specific $(M^z_{j,i}, b^z_{j,i})$ for a polyhedron $P^z_{j,i} \in P^z$ that is a composition of $P^f_j \in P^f$ and $P^g_i \in P_g$ together with corresponding affine transformation $(M^f_j, b^f_j)$ and $(M^g_i, b^g_i)$. We derive the values of a composed affine transformation:
% \begin{align*}
%     z(x) = f(g(x)) &= \\
%     M^f_jg(x) + b^f_j &= \\
%     M^f_j(M^g_ix + b^g_i) + b^f_j &= \\
%     M^f_jM^g_ix + M^f_jb^g_i + b^f_j
% \end{align*}
% From this, we conclude:
% % $$(M^z_{j,i}, b^z_{j,i}) = (M^f_jM^g_i, M^f_jb^g_i + b^f_j)$$
% Correspondingly, this equality was used to compose affine transformations in Coq:
\begin{lstlisting}
Definition compose_affine_functions {in_dim hidden_dim out_dim: nat} 
    (M_f: matrix (T:=R) out_dim hidden_dim) (b_f: colvec out_dim)
    (M_g: matrix (T:=R) hidden_dim in_dim) (b_g: colvec hidden_dim) :=
    (Mmult M_f M_g, Mplus (Mmult M_f b_g) b_f).
\end{lstlisting}
% In this code, \textit{M\_f} and \textit{b\_f} represent $M^f_j$ and $b^f_j$ and similarly, \textit{M\_g} and \textit{b\_g} correspond to $M^g_i$ and $b^g_i$.
% In addition to the presented construction of $P^z$ and $A^z$, we also had to prove the Axiom, which involved proving several properties of individual functions and fixpoints presented. The detailed description of the proof is not in scope in this paper. In addition to the axiom proof, we have formally proven the following theorem:
% \begin{theorem}[Composition correctness]
% Let $f: \mathbb{R}^l \rightarrow \mathbb{R}^n$ and $g: \mathbb{R}^m \rightarrow \mathbb{R}^l$ two PWA functions and $z = f \circ g$, a PWA function as well. For all $x \in R^m$, if the following is satisfied
% \begin{enumerate}
%     \item $x$ is in the domain of $g$, $g(x)$ exists
%     \item $g(x)$ is in the domain of $f$, $f(g(x))$ exists
% \end{enumerate}
% then it holds that
% \begin{enumerate}
%     \item $z(x)$ exists, $x$ is in the domain of $z$
%     \item $z(x) = f(g(x))$
% \end{enumerate}
% \end{theorem}

\paragraph{Univalence of Composition.}
Due to the level of details, the \coq proof for the composed function $z$ satisfying univalence is not included in this paper (see Theorem~\texttt{pwaf\_compat\_univalence}).

\paragraph{Composition Correctness.}
For establishing the correctness of the composition, we proved the following theorem: 
\begin{lstlisting}
Theorem pwaf_compose_correct:
    forall in_dim hid_dim out_dim x f_x g_x 
        (f: PWAF hid_dim out_dim) (g: PWAF in_dim hid_dim),
        in_pwaf_domain g x   -> is_pwaf_value g x g_x ->
        in_pwaf_domain f g_x -> is_pwaf_value f g_x f_x ->
        let fg := pwaf_compose f g in
        in_pwaf_domain fg x /\ is_pwaf_value fg x f_x.
\end{lstlisting}

For one of the lemmas (\texttt{compose\_polyhedra\_subset\_g}) we proved that polyhedra of $g$ are only getting smaller by composing $g$ with $f$ while the borders that are set by polyhedra of $g$ being kept. 

\subsection{Concatenation: Layers of Neural Networks as PWA Functions}
\label{ss:concatenation}

While some neural networks come with each layer being \emph{one} multi-dimensional function, many neural networks feature layers where each neuron is assigned the same lower dimensional function 
independently then applied to each neuron's input. 
Motivated by the transformation of a neural network into a single \pwa function, we introduce a binary operation \emph{concatenation} that constructs a single \pwa function for each \pwa layer of a neural network.
Otherwise, concatenation is interesting due to constructing a multi-dimensional \pwa function being challenging since a user has to define multiple polyhedra with a significant number of constraints. For illustration, we construct a multi-dimensional \relu layer.
 
%\begin{definition}[concatenation]
%Let $f: \mathbb{R}^m \rightarrow \mathbb{R}^n$, $g: \mathbb{R}^k \rightarrow \mathbb{R}^l$, $x_1 \in \mathbb{R}^m$ and $x_2 \in \mathbb{R}^k$. The \emph{concatenation} $ \oplus $ is defined as:
%$$f(x_1) := \begin{bmatrix}
%    f_1(x_1) \\
%    f_2(x_1) \\
%    \dots  \\
%    f_n(x_1)
%\end{bmatrix},\,
%g(x_2) := \begin{bmatrix}
%    g_1(x_2) \\
%    g_2(x_2) \\
%    \dots \\
%    g_l(x_2)
%\end{bmatrix}, 
%$$
%(f \oplus g) (x_1 \oplus x_2) := 
%\begin{bmatrix}
%    f_1(x_1) \\
%    \dots  \\
%    f_n(x_1) \\   
%    g_1(x_2) \\
%    \dots \\
%    g_l(x_2)
%\end{bmatrix}
%$$
%\end{definition}

Concatenation of \pwa functions has to yield an instance of type \texttt{PWAF} since being closed over \pwa functions. Concatenation
is defined as follows:
\begin{definition}[Concatenation]
Let $f: \mathbb{R}^m \rightarrow \mathbb{R}^n$ and $g: \mathbb{R}^k \rightarrow \mathbb{R}^l$. The concatenation $\oplus$ is defined as:
\[
{ (f \oplus g) (\begin{bmatrix}
    x^f \\
    \hline
    x^g
\end{bmatrix}}) := \begin{bmatrix} 
                                  \begin{array}{c}
                                         f(x^f) \\
                                         \hline
                                         g(x^g)
                                  \end{array}
                            \end{bmatrix}
\]
\end{definition}

%\begin{lstlisting}
%Definition pwaf_concat {in_dim1 in_dim2 out_dim1 out_dim2: nat} 
%    (f: PWAF in_dim1 out_dim1) (g: PWAF in_dim2 out_dim2):
%    PWAF (in_dim1 + in_dim2) (out_dim1 + out_dim2) :=
%    mkPLF (in_dim1 + in_dim2) (out_dim1 + out_dim2) (pwaf_concat_body f g) 
%        (pwaf_concat_prop f g).
%\end{lstlisting}

\paragraph{Piecewise Construction of Concatenation.}
Assume some $f,g$, $\Pset^f,\Pset^g, \Aset^f$ and $\Aset^g$ as previously used, and $z = f \oplus g$.
The polyhedra set $\Pset^z$ contains the pairwise joined polyhedra of $\Pset^f$ and $\Pset^g$ but with each constraint of a polyhedron lifted to the dimension of $z$'s domain.
Consider a pair $P^f_i \in \Pset^f$ and $P^g_j \in \Pset^g$. 
For constraints $(c^f_i \cdot x^f \leq b^f_i) \in \C(P^f_i)$ and $(c^g_j \cdot x^g \leq b^g_j) \in \C(P^g_j)$ with $\begin{bmatrix}
    x^f \\
    \hline
    x^g
\end{bmatrix} \in \R^{dim(f) + dim(g)}$, the following higher dimensional constraints are in $\C(P^z_{i,j})$ with $P^z_{i,j} \in \Pset^z$:
$\begin{bmatrix}
    c^f_i \\
    \hline
    0
\end{bmatrix}
\cdot 
\begin{bmatrix}
    x^f \\
    \hline
    x^g
\end{bmatrix}
\leq b^f_i
$ and $\begin{bmatrix}
    0 \\
    \hline
    c^g_j
\end{bmatrix}
\cdot 
\begin{bmatrix}
    x^f \\
    \hline
    x^g
\end{bmatrix}
\leq b^g_j
\text{.}
$
% \noindent\begin{minipage}{.4\linewidth}
% $$\begin{bmatrix}
%     c^f_i \\
%     \hline
%     0
% \end{bmatrix}
% \cdot 
% \begin{bmatrix}
%     x^f \\
%     \hline
%     x^g
% \end{bmatrix}
% \leq b^f_i
% $$
% \end{minipage}%
% \begin{minipage}{.2\linewidth}
% \ \ \ \ \ \ \ and
% \end{minipage}
% \begin{minipage}{.4\linewidth}
% $$\begin{bmatrix}
%     0 \\
%     \hline
%     c^g_j
% \end{bmatrix}
% \cdot 
% \begin{bmatrix}
%     x^f \\
%     \hline
%     x^g
% \end{bmatrix}
% \leq b^g_j
% \text{.}
% $$
% \end{minipage}
Thus, we get
$
\begin{bmatrix}
    x^f \\
    \hline
    x^g
\end{bmatrix}
\in P^z_{i,j}\text{ iff } x^f \in P^f_i\text{ and }x^g \in P^g_j\text{.}$

Hence, the concatenation requires the pairwise join of all polyhedra $\Pset^f$ and $\Pset^g$ each with their constraints lifted to the higher dimension of $z$'s domain:
\begin{lstlisting}
Definition concat_polyhedra {in_dim1 in_dim2: nat}
    (p_f: ConvexPolyhedron in_dim1) (p_g: ConvexPolyhedron in_dim2): 
    ConvexPolyhedron (in_dim1 + in_dim2) :=
    match p_f with | Polyhedron l1 =>
        match p_g with | Polyhedron l2 => 
            Polyhedron (in_dim1 + in_dim2) 
                (extend_lincons_at_bottom l1 (in_dim1 + in_dim2) ++ 
                extend_lincons_on_top l2 (in_dim1 + in_dim2))
    end end.
\end{lstlisting}
 The \coq code uses two functions for insertion of zeros similar to the dimension operations (see Section~\ref{s:preliminaries}).
The corresponding affine function of $P^z_{i,j}$ is then:
 $$(M^z_{i,j}, b^z_{i,j}) := 
 (\begin{bmatrix}
     M^f_i & \vline & 0 \\
     \hline 
     0 & \vline & M^g_j
 \end{bmatrix},
 \begin{bmatrix}
    b^f_i\\
    \hline
    b^g_j
 \end{bmatrix})\text{.}$$
%\begin{lstlisting}
%Definition concat_affine_functions {in_dim1 in_dim2 out_dim1 out_dim2: nat}  
%    (M_f: matrix (T:=R) out_dim1 in_dim1) (b_f: colvec out_dim1)
%    (M_g: matrix (T:=R) out_dim2 in_dim2) (b_g: colvec out_dim2):
%    (matrix (out_dim1 + out_dim2) (in_dim1 + in_dim2) * colvec (out_dim1 + out_dim2))
%    := (block_diag_matrix M_f M_g, colvec_concat b_f b_g).
%\end{lstlisting}

\paragraph{Univalence of Concatenation.}
The technical proof of concatenation being univalent is outside of the scope of this paper (see \texttt{pwaf\_concat\_univalence}).
\paragraph{Concatenation Correctness.}
We proved the correctness of the concatenation:
\begin{lstlisting}
Theorem pwaf_concat_correct:
    forall in_dim1 in_dim2 out_dim1 out_dim2 x1 x2 f_x1 g_x2 
      (f: PWAF in_dim1 out_dim1) (g: PWAF in_dim2 out_dim2),
      in_pwaf_domain f x1 -> is_pwaf_value f x1 f_x1 ->
      in_pwaf_domain g x2 -> is_pwaf_value g x2 g_x2 ->
      let fg   := pwaf_concat f g in
      let x    := colvec_concat x1 x2 in
      let fg_x := colvec_concat f_x1 g_x2 in
      in_pwaf_domain fg x /\ is_pwaf_value fg x fg_x.
\end{lstlisting}
The proof relies on an extensive number of lemmas connecting matrix operations to block matrices and vector reshaping.

\paragraph{Example: \relu Layer.}
Using concatenation, we construct a multi-dimensional \relu layer using one-dimensional \relu (see Section~\ref{ss:nn-model}). 
To construct a \relu layer $\R^n \rightarrow \R^n$, we perform $n$ concatenations of one-dimensional \relu:
\begin{lstlisting}
Fixpoint ReLU_PWAF_helper (in_dim: nat): PWAF in_dim in_dim :=
    match in_dim with
    | 0 => OutputPWAF (in_dim:=0) (out_dim:=0)
    | S n => pwaf_concat ReLU1dPWAF (ReLU_PWAF_helper n)
    end.
\end{lstlisting}

\subsection{Transforming a Neural Network into a \pwa Function}
\label{ss:transformation}

\begin{figure}[!bht]
    \begin{center}
\includegraphics[scale=0.85]{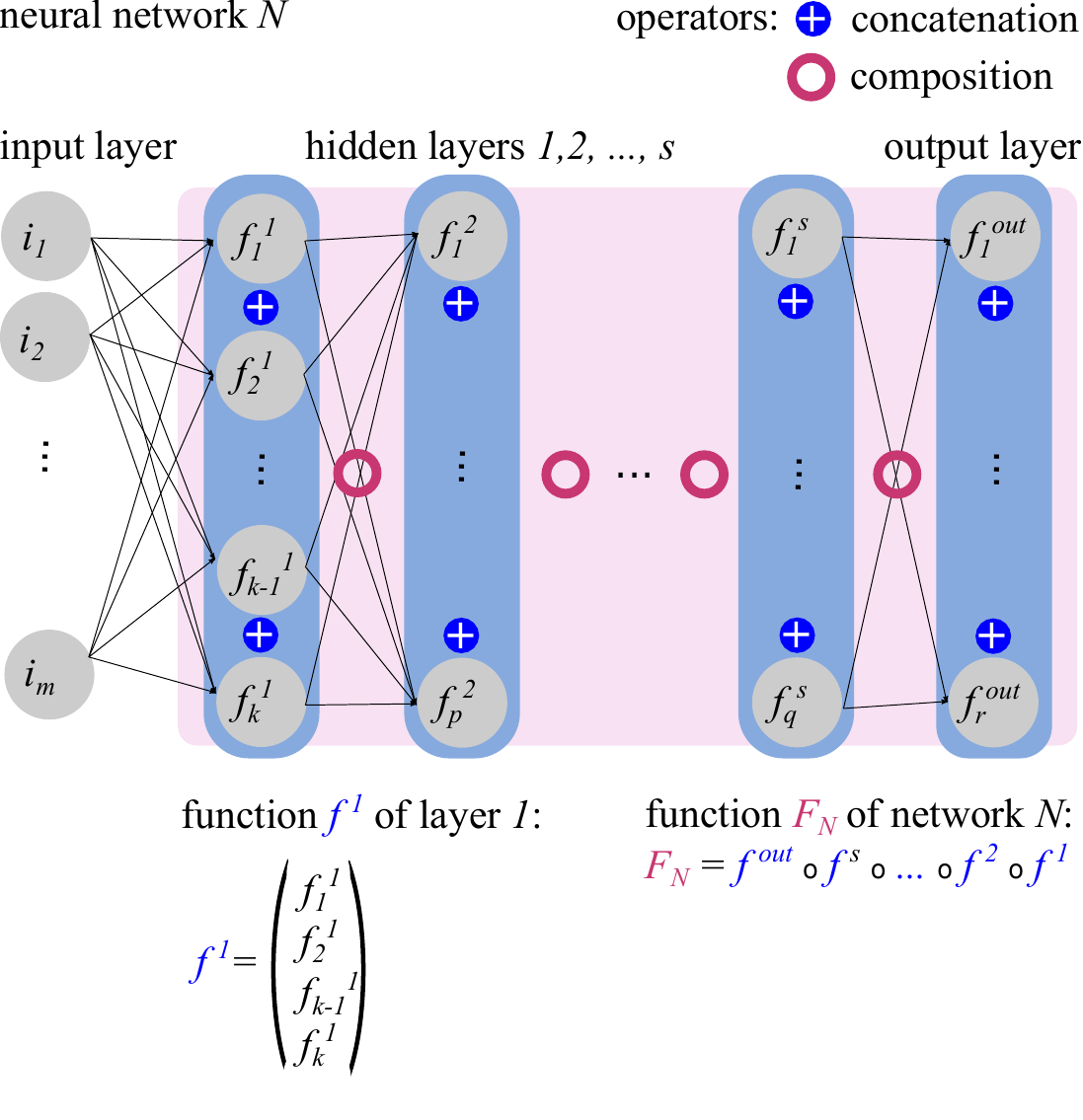}
\end{center}
    \caption{Transformation of a feedforward network $N$ with \pwa activation functions into its representation as a \pwa function $F_N$ by concatenating neuron activation within each layer followed up by composing \pwa layers.}
\label{fig:nn-as-paw-function} 
\end{figure}

Building up on previous efforts, the transformation of a feedforward neural network with \pwa activation functions into a single \pwa function is straightforward.
Using concatenation, we construct multi-dimensional \pwa layers and then compose them to one \pwa function representing the whole neural network.
The transformation is illustrated conceptually in Figure~\ref{fig:nn-as-paw-function}. 

The transformation in \coq simply fails when applied to hidden layers that are non-\pwa:
\begin{lstlisting}
Fixpoint transform_nn_to_pwaf {in_dim out_dim: nat}
    (nn: NNSequential (input_dim := in_dim) (output_dim := out_dim)) 
    : option (PWAF in_dim out_dim) :=
    match nn with
        | NNOutput => Some (OutputPWAF)
        | NNPlainLayer _ _ _ => None
        | NNUnknownLayer _ _ => None
        | NNPWALayer _ pwaf next => 
            match transform_nn_to_pwaf next with
            | Some next_pwaf => Some (pwaf_compose next_pwaf pwaf)
            | None => None
    end end.
\end{lstlisting}
%\filbreak

\paragraph{Correctness of Transformation.}
For this transformation, we have also proven the following theorem in \coq to establish its correctness:
\begin{lstlisting}
Theorem transform_nn_to_pwaf_correct:
    forall in_dim out_dim (x: colvec in_dim) (f_x: colvec out_dim) nn nn_pwaf,
        Some nn_pwaf = transform_nn_to_pwaf_correct nn ->
        in_pwaf_domain nn_pwaf x ->
        is_pwaf_value nn_pwaf x f_x <-> nn_eval nn x = Some f_x.
\end{lstlisting}
For a neural network $\N$ and its transformed \pwa function $f_{\N}$, the theorem states that for all inputs $x \in dom(f_{\N})$ holds $f_{\N}(x) = \N(x)$.
The proof of this theorem relies on several relatively simple properties of the composition.
Note that for $dom(f_{\N})=\emptyset$ the theorem trivially holds, and in fact an additional proof is required for $f_{\N}$'s polyhedra being a subdivision of $dom(\N)$ (i.e. $dom(f_{\N}(x)) = dom(\N(x))$).
%Sounds great but I would leave it out on this paper. I think what we want is to prove the property of a subdivision to cover a given space rather that totality.
%Further, the assumption of $x$ being in domain of $f_{\N}$ may complicate the use of the theorem. 
%We observed that all activation functions in our Coq library are total (defined on the entire $\mathbb{R}^m$) and 
%we expect that additional lemmas and an extended type of \textit{total \pwa functions} may simplify the domain proof significantly or even remove this assumption completely.
%\filbreak

\paragraph{On the Representation of a Neural Network as a \pwa Function.}
The main benefit of having a \pwa function obtained from neural network lies in the option to use simple-to-implement encodings of \pwa functions for different solvers, e.g. \coq's tactic \texttt{lra} or \textsc{MILP}/\textsc{SMT} solvers. Hence, this representation paves the way for proof automation when stating theorems about the input-output relation of a network in \coq. 

Furthermore, a representation as a \pwa function moves the structural complexity of a neural network to the polyhedral subdivision of the \pwa function. 
This is interesting since local search can be applied to the set of polyhedra for reasoning about reachability properties in neural networks~\cite{rpm2021}. 
%These algorithms can also be verified in \coq.
Furthermore, one may estimate the size of a \pwa function's polyhedral subdivision for different architectures 
of neural network~\cite{montufar2014onthenumber}.

%Additionally, a \pwa function is useful when it comes to end-to-end verification where a model of the environment in which the neural networks operates in is taking into account.
%ybrid systems are expressable as 
%\todo[color=yellow]{Kontinuierliche hybride Systeme sind durch pwa nur approximierbar, aber prinzipiell kann man vielleicht einige Klassen diskreter dynamischen Systeme (auch switching) als pwa darstellen (x(n+1) = A*x(n) + b). Da es keine generelle Eigenschaft der hybriden Systeme ist, sollen wir diese vielleicht nicht nennen?}
%a system model can be composed out of a neural networks and a model of its training environment (often a simulation) that can be expressed or approximated by a \pwa function. This approach may result in a system-level verification framework inside \coq. 

%% file: sections/discussion.tex
\section{Discussion}
\label{s:discussion}

We were working towards neural network verification in \coq
with a verified transformation from a network to a \pwa function being the main contribution.

\paragraph{Summary.}
We presented the first formalization of \pwa activation functions 
for an interactive theorem prover tailored to verifying neural networks with \coq.
For our constructive formalization, we used a \pwa function's polyhedral subdivision
due to the numerous efficient algorithms working on polyhedra. 
Our class of \pwa functions is on-purpose restricted  to suit linear programming 
by using non-strict constraints
and to fit \textsc{SMT}/\textsc{MILP} solvers by employing finitely many polyhedra.
Using \coquelicot's reals, we enabled reasoning about gradients and support \coq's tactic \texttt{lra}.
With \relu, we constructed one of the most popular activation functions.
We presented a verified transformation from a neural network to its representation as a \pwa function
enabling encodings for proof automation for theorems about the input-output relation.
To this end, we devised a sequential model of neural networks,
and introduced two verified binary operation on \pwa functions -- usual function composition together
with an operator to construct a \pwa function for each layer.

\paragraph{Future Work.}
Since the main benefit of having a \pwa function obtained from neural network 
lies in the many available encodings~\cite{10.5555/3327345.3327388, Ehlers2017FormalVO} 
targeting different solvers, we envision encodings for our network model.
These encodings have to be adapted to the verification within \coq
with our starting point being the tactic \texttt{lra} -- a \coq-native decision procedure for linear real arithmetic.

Moreover, moving the structural complexity of a neural network to the polyhedral subdivision of a \pwa function,
opens up on investigating algorithms working on polyhedra for proof automation with
our main candidate being local search on polyhedra for reasoning about reachability properties in neural networks~\cite{rpm2021}. 

Further, for our model of neural networks, we intend a library of \pwa activation functions 
with proof automation to ease construction.
We also plan on a generic graph-based model for neural networks in \coq but as argued, 
we expect the sequential model to stay the mean of choice for feedforward networks.
Additionally, since tensors are used in machine learning to incorporate complex mathematical operations,
we aim to integrate a formalization of tensors tailored to neural network verification.